\documentclass[conference]{IEEEtran}
\IEEEoverridecommandlockouts
\usepackage{cite}
\usepackage{amsmath,amssymb,amsfonts}
\usepackage{algorithmic}
\usepackage{graphicx}
\usepackage{textcomp}
\usepackage{xcolor}

\usepackage{caption}
\usepackage{subcaption}
\usepackage{float}

\def\BibTeX{{\rm B\kern-.05em{\sc i\kern-.025em b}\kern-.08em
    T\kern-.1667em\lower.7ex\hbox{E}\kern-.125emX}}
\begin{document}

\title{KarNet: An Efficient Boolean Function Simplifier\\
 }

\author{\IEEEauthorblockN{Shanka Subhra Mondal, Abhilash Nandy, Ritesh Agrawal, Debashis Sen}
\IEEEauthorblockA{\textit{Department of Electronics and Electrical Communication Engineering, Indian Institute of Technology}\\
Kharagpur, India \\
shankasubhra@iitkgp.ac.in, raj12345@iitkgp.ac.in, riteshagrawal@iitkgp.ac.in, dsen@ece.iitkgp.ac.in}
}

\maketitle

\begin{abstract}
Many approaches such as Quine-McCluskey algorithm, Karnaugh map solving, Petrick's method and McBoole's method have been devised to simplify Boolean expressions in order to optimize hardware implementation of digital circuits. However, the algorithmic implementations of these methods are hard-coded and also their computation time is proportional to the number of minterms involved in the expression. In this paper, we propose \emph{KarNet}, where the ability of Convolutional Neural Networks to model relationships between various cell locations and values by capturing spatial dependencies is exploited to solve Karnaugh maps. In order to do so, a Karnaugh map is represented as an image signal, where each cell is considered as a pixel.  Experimental results show that the computation time of \emph{KarNet} is independent of the number of minterms and is of the order of one-hundredth to one-tenth that of the rule-based methods. \emph{KarNet} being a learned system, is found to achieve nearly hundred percent accuracy, precision and recall. We train \emph{KarNet} to solve four variable Karnaugh maps and also show that a similar method can be applied on Karnaugh maps with more variables. Finally, we show a way to build a fully accurate and computationally fast system using \emph{KarNet}.
\end{abstract}


\section{Introduction}
\label{sec:intro}
Numerous methods have been employed over the years in order to obtain simplified Boolean expressions of variables from sums of minterms or product of maxterms. For instance, Karnaugh map solving \cite{b1}, Quine McCluskey algorithm \cite{b2} are some of the popular ones in this domain. These methods are based on a specific set of rules, and hence are hard-coded. Further, these problems have non polynomial (NP)-hard complexity \cite{b15}.

\begin{figure}[h]
\centering
    \begin{subfigure}[b]{0.19\textwidth}
        \includegraphics[width=\linewidth]{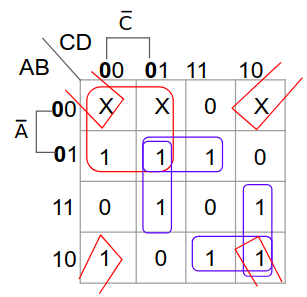}
        \caption{Sum of Products}
        \label{fig:siamese_2a}
    
    \end{subfigure}%
    \begin{subfigure}[b]{0.23\textwidth}
        \includegraphics[width=\linewidth]{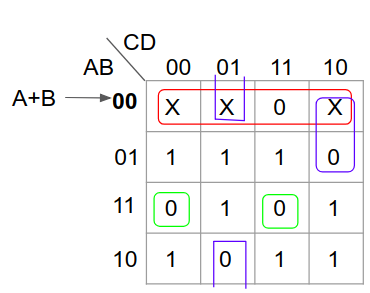}
        \caption{Product of sums}
        \label{fig:siamese_2b}
    \end{subfigure}%
\caption{Karnaugh maps}
\label{fig:siamese_2}
\end{figure}
\begin{figure}[H]
\centering
    \includegraphics[width=0.8\linewidth]{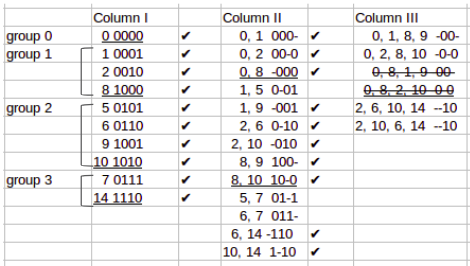}
    \caption{Table of prime implicants}
    \label{fig:siamese_pi}
\end{figure}
\begin{figure}[h]
\centering
    \includegraphics[scale=0.5]{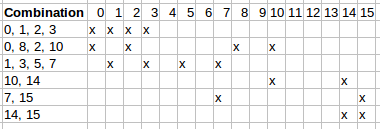}
    \caption{Quine McCluskey Method}
    \label{fig:siamese_qm}
\end{figure}

One of the famous rule-based methods is the Karnaugh map solving \cite{b1} (abbreviated as K-map), in which, the minterms (or maxterms, as is the case) are marked as 1s (or 0s) in a square grid and don't cares as Xs whose size is (number of variables) x (number of variables) and the cells are ordered according to the gray code. Other entries of the grid are marked as 0s (or 1s). An optimal grouping of the 1s (or 0s) in order to minimize the original expression, is shown in Fig. \ref{fig:siamese_2}. 
Another popular method is the Quine-McCluskey Algorithm \cite{b2}. First, the prime implicants of the given function are found, and then, they are listed in the form of a prime implicant chart, as in Fig. \ref{fig:siamese_pi} and Fig. \ref{fig:siamese_qm}. Prime implicant is an implicant that cannot be covered by a more general implicant, where an implicant is a covering of one or more minterms in the sum of products form of a Boolean function. However, on the downside, the time complexity of this algorithm increases exponentially as the number of variables increase. 


\begin{figure}[H]
    \centering
    \includegraphics[scale = 0.20]{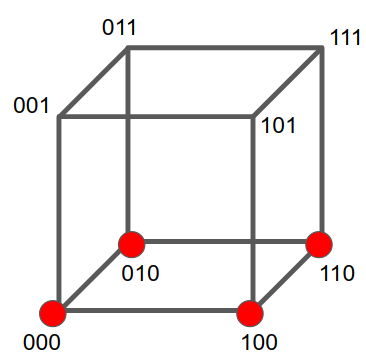}
    \caption{Boolean functions represented as a cube's vertices}
    \label{fig:siamese_cube}
\end{figure}

Petrick's method \cite{b3} is also a rule-based method based on the minimization of sum-of-products from the prime implicant chart. McBoole Method \cite{b4} uses efficient graph partitioning techniques to minimize the Boolean functions, where the Boolean function to be minimized is represented in the form of list of cubes, as shown in Fig.  \ref{fig:siamese_cube}.

\begin{figure}[h]
    \centering
    \includegraphics[scale=0.21]{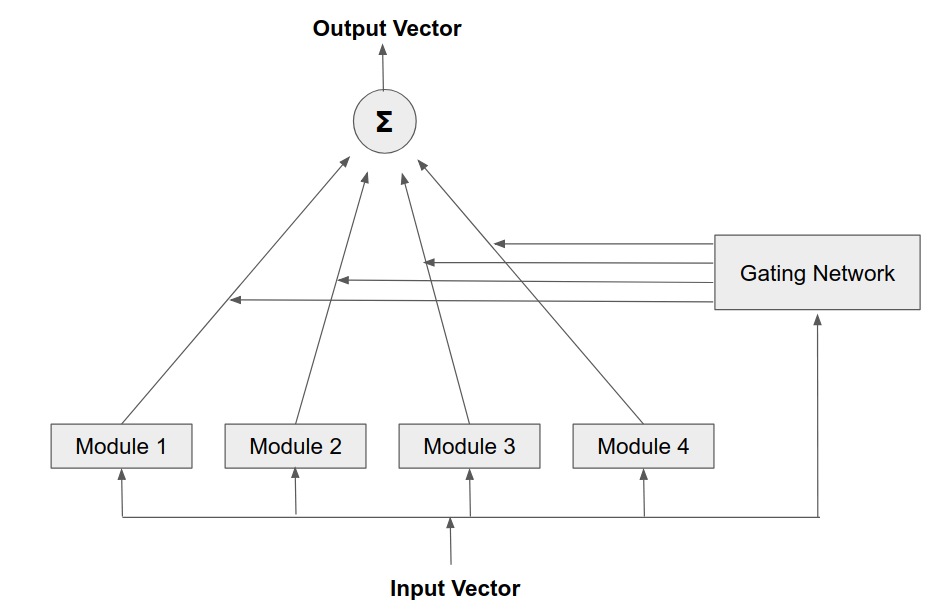}
    \caption{Architecture of Modular Neural Nets (MNNs)}
    \label{fig:MNN}
\end{figure}

Recent works such as \cite{b14} include minimization of Boolean functions using a repetitive block diagram without using any auxiliary objects and achieves lower cost of development and computation compared to Karnaugh maps. In \cite{b16} an efficient realization of deep neural networks in terms of computation resource consumption, memory cost has been proposed using Boolean logic minimization. Neural Networks also have been used in the past to address Boolean logic simplification. For instance,~\cite{b5} used modular neural nets in order to build an efficient learning architecture for Boolean algebra. Modular Neural Nets divide the task into various sub-tasks, each task being executed by a simple neural network. In order to control the weights of each of the neural net's outputs, there is a gating network, which controls the training patterns of the sub-networks, as shown in Fig.~\ref{fig:MNN}.

In~\cite{b6}, a neural network is used to select a subset of prime implicants, which would be able to satisfy the function to be simplified for all possible input values with a minimal overall cost, where cost for a prime implicant could mean the number of literals associated with it.

All the rule-based methods described above and~\cite{b6} have a computation time proportional to the number of minterms. Further, the method described in~\cite{b5} has many parameters to be learned, since it involves multiple neural networks, thus increasing computational time.

In this paper, we propose \emph{KarNet}, which learns Boolean simplification by itself. We give a Karnaugh map as an input, represented as an image signal, to a convolutional neural network (CNN). The CNN is designed using appropriate filters and padding, which maps the input to its simplified Boolean expression obtained as the output. \emph{KarNet} has a computation time that is independent of the the number of minterms, which is much lower than the algorithmic methods. Moreover, we show that it achieves nearly hundred percent accuracy, precision and recall on all possible input four variable Karnaugh maps. We further show that \emph{KarNet} can perform in cases of higher number of variables as well. Finally, we suggest a way to built a hundred percent accurate (like the algorithmic methods) and computationally fast system with \emph{KarNet}.

The paper is organized as follows. The motivation behind the proposed solution is presented in Section II, and the approach is explained in Section III. The baselines and comparative experimental results are detailed in Sec. IV. Sec. V concludes the article.



\section{Motivation}

Convolutional neural networks (abbreviated as CNNs) are widely used for image classification \cite{b7}, localization \cite{b12} and segmentation tasks \cite{b13}. CNNs consist of various filters arranged in a hierarchical fashion doing convolution operations on the pixels of an input image. The filter coefficients are learned by the network from available training data. By repeated application of the convolutional filters, the network is able to extract learned features (filtered outputs) from the image data, with the filters at the beginning of the architectural hierarchy capturing the basic features and the filters in the latter part of the hierarchy capturing complex features. Thus the relationship between various pixel locations and values are captured along the said hierarchy. Inspired by the use of neural networks in the domain of Boolean algebra \cite{b5,b6,b16}, we present a CNN based Boolean simplification method (\emph{KarNet}), where the network is suited to learn the class mappings of input Karnaugh maps represented as image signals to get the product terms of the sum-of-products in simplified form.

\section{Proposed Method}\label{prop}

Since, the method of grouping the cells in a Karnaugh map depends on the values and the locations of the pixels, we use a CNN based architecture, which would learn the pixel-wise relationships (treating each cell of a Karnaugh map as a pixel), and do the grouping of cells into product terms. Further, using locally connected CNNs instead of generic multi layer perceptrons, would require much less number of learnable parameters, and reducing memory requirement.



\begin{figure}[h]
    \centering
    \includegraphics[width=0.11\textwidth]{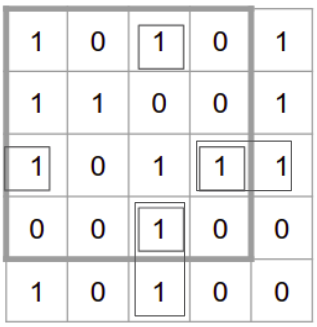}
    \caption{One level of padding - on the bottom and on the right.}
    \label{fig:padding}
\end{figure}

\begin{figure*}
\centering
    \includegraphics[scale = 0.45]{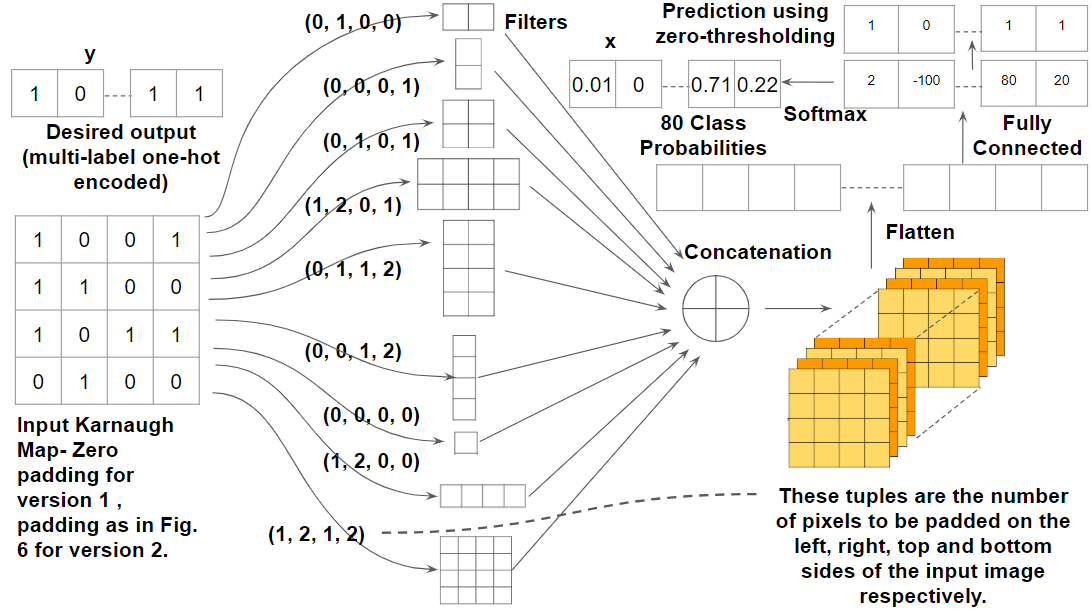}
    \caption{Schematic of \emph{KarNet} Architecture.}
    \label{fig:propnet}
\end{figure*}

We first consider four variable K-maps in order to solve the problem of finding the minimum sum of products. As in Fig. \ref{fig:propnet}, we use the Karnaugh map as a 4x4 image along with some padding as input . The desired output is provided in an one-hot encoded manner, in which, out of all possible product terms (say $AB$, $ACD$ etc., $A$, $B$, $C$ and $D$ being the variables) as the total number of output classes, the desired product terms are marked as 1. 
Since, solving Karnaugh maps involves making rectangular groups of $1$s, each group having number of terms in powers of 2, we take inspiration from the same. The architecture that we propose involves a padded $4$x$4$ Karnaugh map as input, on which, $8$ different sizes of convolutional filters are applied. The filter sizes are $1$x$1$, $1$x$2$, $2$x$1$, $1$x$4$, $2$x$2$, $4$x$1$, $2$x$4$, $4$x$2$ and $4$x$4$, with appropriate padding applied to the input according to the filter used, so that, all convolutional filters give output feature maps having shape same as that of the input. For one version of the proposed method (\emph{KarNet}-1), we apply zero-padding. For another version (\emph{KarNet}-2), padding of zeros is applied in addition to padding of the first row below and padding of the first column on the right, so as to learn the edge rectangular groupings in a more efficient manner, as shown in Fig.~\ref{fig:padding}.

All feature maps obtained are then concatenated, flattened, and passed through two fully connected layers of neurons, and finally, the softmax activation function \cite{b8} is applied to estimate the probabilities of each of the classes as shown in Fig.~\ref{fig:propnet}.

For training \emph{KarNet}, multi-label soft margin loss function is used. This loss function in  (\ref{multi}) optimizes a multi-label one-versus-all loss based on max-entropy between the predicted output and the desired output~\cite{b9}.

\begin{eqnarray}
\label{multi}
loss(\textbf{x},\textbf{y}) = -\sum_{i = 0}^{N - 1}\Biggl\{y[i]*log((1 + exp(-x[i]))^{-1}) +\nonumber \\ (1- y[i]) * log\Biggl(\frac{exp(-x[i])}{(1 + exp(-x[i]))}\Biggl)\Biggl\}
\end{eqnarray}

In the above expression, $\textbf{x}$ refers to the vector of predicted class probabilities, $\textbf{y}$ refers to the one-hot encoded vector representing the class(es) to which the input belongs to, and $N$ is the number of classes used for classification purpose. Final prediction from the output of the network is done by zero thresholding the intermediate output of the network just before applying the softmax function. The neurons for which the value is greater than zero correspond to the predicted classes which are then mapped to the product terms to obtain the minimum Boolean representation of the function in the sum of products form.

\section{Experimental Results}

\subsection{Training Data}\label{data_extraction}
Here we elaborate on the data used to train \emph{KarNet}. There are $2^{16}$ different K-maps possible for 4 variables, out of which, two of the Karnaugh maps are trivial, one having all $1$s, and the other having all $0$s and the total number of classes are 80. In order to extract the K-maps, we have used Quine-McCluskey Method, taking the minterms as the input, and giving as output the simplified equation. So, according to the minterms in the input, the corresponding positions would be 1 in the K-map, and the rest of the positions would be 0. The simplified equation is converted to one-hot encoded vectors, as has already been discussed in Section~\ref{prop}.

\subsection{Baseline algorithms}
We consider a few baseline algorithms for comparison with \emph{KarNet}, which are listed here. The input image signal is of size closer to the MNIST handwritten image data \cite{b17}, which is of size $32$x$32$. Hence, we chose some of the state-of-the-art models of the MNIST data and some additional  baselines to highlight some important factors which is explained in  Section~\ref{results} to benchmark the proposed method. The neural network architectures (earlier used on \cite{b17}) are as follows:

\textbf{A1} - Conv2d((3, 3 x 3), stride=(1, 1), padding=(1, 1)) -- ReLU() -- Bilinear\_Interpolation() -- Fully \\Connected(in\_features=192, out\_features=80).

 \textbf{A2} - Conv2d((3, 3 x 3), stride=(1, 1), padding=(1, 1)) -- ReLU() -- UpConv2d((3, 2 x 2), stride=(2, 2)) -- Fully Connected(in\_features=192, out\_features=80).
 
 \textbf{A3} - Conv2d((12, 3 x 3), stride=(1, 1), padding=(1, 1)) -- ReLU() -- UpConv2d((72, 2 x 2), stride=(2, 2)) -- Fully Connected(in\_features=4608, out\_features=80).
 

\textbf{A4}- This is based on LeNet architecture~\cite{b11}, in which, the input Karnaugh map is deconvoluted in order to increase the feature map's size to $32$x$32$, following which, we apply the LeNet architecture, the only difference being that, in the last layer, we use the number of neurons equal to the number of classes in our case, which is 80.

\textbf{A5}- Deep columnar CNN architecture \cite{b10} with the input Karnaugh map deconvoluted to $28$x$28$. 

In these baselines, the minimum representation of the Boolean function in the sum of products form is obtained in a similar fashion like the proposed method by zero thresholding the intermediate output and mapping the positive activations to the product terms.

\subsection{Training Parameters and Hyperparameters}

We did a train:validation:test split as 70:20:10, stratified on the basis of the number of classes corresponding to each input. Adam optimizer with a learning rate of $5$x$10^{-4}$ is used. Training is done on batches, with a batch size of $64$, and is continued for $600$ epochs. The model used for testing is the one saved at the epoch with the highest validation accuracy.

\subsection{Evaluation and Results}\label{results}

We used  accuracy, average precision, average recall and computation time on the test set as the evaluation metrics. For the architectures A1, A2, A3, A4, A5, and the two versions of the proposed method discussed above, Table~\ref{fig:tab} contains evaluation results.
\begin{table}[H]
\centering
\caption{Comparison of various neural network based methods. Best performance metric is indicated in bold. }
\label{fig:tab}
\begin{tabular}{|c|c|c|c|c|c|}
\hline
\scriptsize \textbf{Arch.}&\begin{tabular}[c]{@{}c@{}}\scriptsize \textbf{Accuracy}\\ \end{tabular} & \begin{tabular}[c]{@{}c@{}}\scriptsize \textbf{Avg}  \\\scriptsize \textbf{Precision}\end{tabular} & \begin{tabular}[c]{@{}c@{}}\scriptsize \textbf{Avg}\\\scriptsize \textbf{Recall}\end{tabular} & \begin{tabular}[c]{@{}c@{}}\scriptsize \textbf{Avg}\\\scriptsize \textbf{F1-Score}\end{tabular} & \begin{tabular}[c]{@{}c@{}}\scriptsize \textbf{No. of}\\ \scriptsize \textbf{Parameters}\end{tabular} \\ \hline
\scriptsize A1    & \scriptsize 40.79\%                                                & \scriptsize 86.78\%                                             & \scriptsize 86.35\% &   \scriptsize 86.56\%                                         & \scriptsize 15,470                                                               \\ \hline
\scriptsize A2    & \scriptsize 62.22\%                                                & \scriptsize 90.85\%                                             & \scriptsize 90.43\%  & \scriptsize 90.64\%                                           & \scriptsize 15,509                                                              \\ \hline
\scriptsize A3    & \scriptsize 74.14\%                                                & \scriptsize 95.78\%                                              & \scriptsize 96.23\%     &   \scriptsize 96.00\%                                      & \scriptsize 3,72,368                                                               \\ \hline
\scriptsize A4 
    & \scriptsize 89.08\%                                                & \scriptsize 98.25\%                                              & \scriptsize 98.57\%
     & \scriptsize 98.40\% 
    & \scriptsize 68,151                                                               \\ \hline
\scriptsize A5 
& \scriptsize 93.72\%                                                &\scriptsize 98.63\%                                              & \scriptsize 98.53\% 
& \scriptsize 98.58\%                                              & \scriptsize 19,60,066                                                               \\ \hline
\scriptsize \emph{KarNet}-1    & \scriptsize 92.97\%                     & \scriptsize 98.97\%                                              & \scriptsize 98.99\%  
 & \scriptsize 98.98\% 
& \scriptsize 4,53,584                                                               \\ \hline
\scriptsize \emph{KarNet}-2     & \scriptsize \textbf{93.84\%}                                                & \scriptsize \textbf{99.02\%}                                              & \scriptsize \textbf{99.04\%} 
 & \scriptsize \textbf{99.03\%} 
 & \scriptsize 4,53,584                                                               \\ \hline
\end{tabular}
\end{table}

As can be seen from Table~\ref{fig:tab}, \emph{KarNet} performs better than all the baselines. This is because in \emph{KarNet},  filter sizes are such that they correspond to the shapes of the rectangular groups of $1$s which is in accordance with the method of Karnaugh map solving. Among the two versions of the  proposed method, the second one performs better than the first one, which may be due to the padding in the second method that allows the filters to learn even groups of $1$s across opposite edges.

In the baseline algorithms, having the deconvolutional layer in A2 instead of having a bi-linear interpolation function in A1 gives better results, as due to the introduction of deconvolutional layer, the total number of trainable parameters increases, thus being able to learn the complex mapping function more effectively. However, DC-CNN and LeNet baselines performs much better than the other 3 baselines. For LeNet baseline, this might be due to the number of features increasing by 64 times initially using a deconvolutional layer. DC-CNN has the highest number of parameters among all the baselines, and this might be the reason for it performing the best among all baselines.

\textbf{Generalizability of \emph{KarNet}} -\\
\emph{K-maps with higher number of variables:}
Here, we show that \emph{KarNet} is applicable to cases of K-maps with higher number of variables. As an example, without loss of generality, we applied a \emph{KarNet} on K-maps having 5 variables. For this demonstration, we used a part of the $2^{32} - 2$ K-maps data, taking $51,05,504$ K-maps, having $115$ classes. The sampling of these K-maps was done in a stratified fashion, based on the number of classes to which each input belonged to. An accuracy of $92.34\%$, precision of $98.32\%$, recall of $97.82\%$ and F1 score of $98.07\%$ was achieved using \emph{KarNet}-1, whereas the best baseline, A5 gave an accuracy of $90.25\%$, precision of $98.26\%$, recall of $98.25\%$ and F1 score of $98.25\%$. This shows that \emph{KarNet} can be trained to solve K-maps with higher number of variables achieving accuracies close to that obtained for the four variable case in Table~\ref{fig:tab}.

\emph{K-maps with don't care:}
 Here, we show that \emph{KarNet} is applicable to cases of K-maps with don't care cells. We extracted all possible 4 variable Karnaugh maps with the help of Quine Mccluskey Method for don't care terms, which had at least one don't care cell and at least one cell with value $1$. There are $42850116$ such K-maps in total, with the total number of classes being $80$. In order to convert the K-maps to images, the don't care cell was filled with $0.5$, since, it had an equal probability of being occupied by either $0$ or $1$. Performing stratified sampling over all samples based on the number of classes of each input, we sampled $800000$ inputs. Subsequently, following the same train:validation:test split, optimizer and learning rate, a batch size of $4096$, and applying \textit{KarNet}-1 on the input, we achieved an accuracy of $85.58\%$, precision of $98.14\%$, recall of $98.32\%$, and a F1 score of $98.23\%$ on the test set.
 
 The accuracy in case of five variables and don't care K-maps can be increased by using more data for training, but it should be noted that in both the generalization cases, even by using only a small  fraction of relevant data, more than 90\% accuracy is achieved on the overall data.

    
    
    



\begin{figure}[h]
    \centering
    \includegraphics[scale = 0.45]{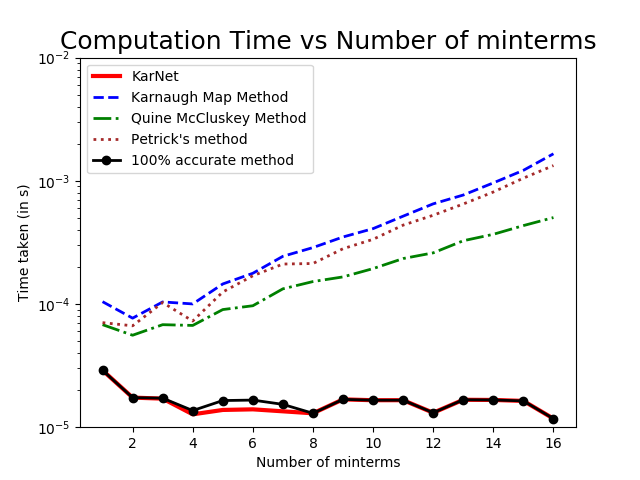}
    \caption{Time taken to solve an example vs. number of minterms for various methods}
    \label{fig:Times}
\end{figure}

\textbf{Computation time} - To highlight our most important contribution, we present here the time to process one set of given minterms to get the reduced Boolean expression for four methods - Karnaugh map solving method, Quine McCluskey Method, Petrick's Method, and \emph{KarNet}. This was performed for varying number of minterms in the input. The time taken is plotted against the number of minterms, as shown in Fig.~\ref{fig:Times}.

This graph suggests two important things - one being that, the computation time of K-map, Quine McCluskey and Petrick's methods are almost 10-100 times higher than \emph{KarNet}, and the other being that, with an increase in the number of minterms, computation time of the rule-based methods increases, while for \emph{KarNet} it is nearly independent of the number of minterms. The time of inference per input for \emph{KarNet} is independent of the number of minterms. This is because \emph{KarNet} treats each input to be simplified as an image, and the feed-forward time is independent of the pixel values of the input. Whereas, for the Karnaugh map solving method, the time taken is proportional to the number of groups of $1$s formed in K-map representation, which is in turn proportional to the number of minterms. For the Quine McCluskey and Petrick's methods, all prime implicants are to be listed and minimal subset satisfying the truth table corresponding to the minterms of the input is to be found, which is proportional to the number of minterms present in the input. Hence, \emph{KarNet} clearly has much lesser computation time when compared with conventional rule-based methods of Boolean function minimization.

\textbf{100\% Accurate Method (using a combination of \emph{KarNet} and Quine McCluskey Method)} - Many applications necessitate that Boolean function simplifier are 100\% accurate, and here we suggest a way to achieve that using our computationally fast \emph{KarNet}. The results of the overall accuracy obtained on inference performed on the whole data corresponding to 4x4 Karnaugh maps are tabulated in Table \ref{tab2}. For the part of the data used for $5$ variable and don't care K-maps, the overall accuracies are $97.33\%$ and $90.2\%$, respectively.
\vspace{-0.2cm}
\begin{table}[H]
\centering
\caption{Accuracy on all 4x4 K-maps}
\label{tab2}
\begin{tabular}{|c|c|c|c|c|c|}
\hline
\textbf{Arch.}            & A3      & A4      & A5      & \emph{KarNet}-1 & \emph{KarNet}-2          \\ \hline
\textbf{Overall Acc.} & 74.10\% & 94.84\% & 96.59\% & 97.09\% & \textbf{98.09\%} \\ \hline
\end{tabular}
\end{table}
\vspace{-0.2cm}
Now for the $4$ variable case, if we consider \emph{KarNet}-2, of $65534$ K-maps, $1250$ K-maps have been predicted wrong. Hence, we can build a $100\%$ accurate Boolean minimization engine having very less computation time by using Quine McCluskey Method to predict the outputs where \emph{KarNet}-2 would be inaccurate (stored as a prior knowledge) and use \emph{KarNet}-2 on the rest of the samples. The computation time as a function of the number of minterms for this method has been plotted in Fig. \ref{fig:Times}, which is just marginally more than that of the \emph{KarNet}.

\section{Conclusion}
Through our proposal \emph{KarNet}, we have shown that the  simplification of Boolean expressions can be performed with the help of neural networks like CNNs with increased efficiency. The computation times of the rule-based methods like Quine-McCluskey and solving K-maps are based on the number of comparisons and are found to increase exponentially with the number of minterms. On the other hand, it is found that more number of minterms does not affect the computation time in case of \emph{KarNet}.
We get close to 100\% accuracy in case of four variable K-maps and with a similar accuracy using a very small fraction of the data in case of five variable K-maps we demonstrate generalizability of \emph{KarNet}. Applicability of \emph{KarNet} on K-maps with don't care conditions is also shown. A fully accurate and efficient system is demonstrated using a combination of \emph{KarNet} and Quine McCluskey method. Given the existing potential, the work shall be extended in future, to solve K-maps with larger number of variables, so as to achieve 100\% accuracy within a significantly less computation time.

\vspace{12pt}
\color{red}

\end{document}